\documentclass[letterpaper,10pt,journal,twoside]{IEEEtran}

\newcommand{\papertitle}{RAIWAY: A Large-Wheeled Bipedal Robot with Contact-Informed Blind Locomotion}

\newcommand{\RALfirstauthorsurname}{NAM}
\newcommand{\RALshorttitle}{\papertitle}

\usepackage{graphicx}
\usepackage{svg}
\usepackage{amsmath, amssymb}
\usepackage{array}
\usepackage{booktabs}
\usepackage[caption=false, font=footnotesize]{subfig}
\usepackage{stfloats}
\usepackage{xcolor}
\usepackage{float}
\usepackage{needspace}
\usepackage{makecell}
\usepackage{hyperref}
\usepackage{cite}
\usepackage{bbm}

\title{\papertitle}

\author{Kyungmin Nam, Jiseong Lee, Donghyuk Choi, Jeongsoo Park, Woojin Kim, and Jemin Hwangbo%
\thanks{%
Received 23 June 2026; accepted 18 October 2026.
This article was recommended for publication by Editor L. Pallottino upon evaluation of the reviewers' comments.
This research was financially supported by the Institute of Civil Military Technology Cooperation funded by the Defense Acquisition Program Administration and Ministry of Trade, Industry and Energy of Korean government under grant No. UM22207RD2. \textit{(Corresponding author: Jemin Hwangbo.)}

\par
The authors are with Robotics and Artificial Intelligence Lab, Korea Advanced Institute of Science \& Technology (KAIST), Daejeon 34141, South Korea (e-mail: twinhk@kaist.ac.kr; ssuni1002016@kaist.ac.kr; cay879@kaist.ac.kr; integerpark@kaist.ac.kr; woojin0624@kaist.ac.kr; jhwangbo@kaist.ac.kr).
}%
}

\begin{document}

\bstctlcite{IEEEexample:BSTcontrol}

\maketitle
% \thispagestyle{empty}
% \pagestyle{empty}

%%%%%%%%%%%%%%%%%%%%%%%%%%%%%%%%%%%%%%%%%%%%%%%%%%%%%%%%%%%%%%%%%%%%%%%%%%%%%%%%
\begin{abstract}
We present RAIWAY, an in-house wheeled bipedal robot with 20-inch wheels, together with a contact-informed control approach for rolling locomotion across irregular and discontinuous terrain. Such locomotion is fundamentally contact-dominated, governed by local wheel--terrain interaction that is impractical to instrument directly and depends on more than terrain geometry alone. The proposed framework exploits the contact information available in simulation, distilling multi-contact events into a latent representation. At deployment, the controller infers this representation online from proprioception, requiring no contact sensing. In real-world experiments, the proposed system traverses diverse environments including steep slopes, gravel beds, studded surfaces, and rocky terrain, and demonstrates continuous ascent on the tested outdoor staircases with riser heights up to 15~cm. On our platform, the proposed controller further outperforms baselines without the contact-informed mechanism across challenging indoor and outdoor settings. The video can be found at https://youtu.be/Ch0divLg64w.
% TODO: change URL
\end{abstract}

\begin{IEEEkeywords}
Wheeled Robots, Reinforcement Learning, Sim-to-Real Transfer
\end{IEEEkeywords}

%%%%%%%%%%%%%%%%%%%%%%%%%%%%%%%%%%%%%%%%%%%%%%%%%%%%%%%%%%%%%%%%%%%%%%%%%%%%%%%%
\section{INTRODUCTION}

\IEEEPARstart{L}{egged} robots have demonstrated remarkable robustness by utilizing articulated legs to traverse uneven terrains and overcome complex obstacles~\cite{Hwangbo2,Hwangbo3,KHJ}. Wheeled systems, in contrast, provide high locomotion efficiency and speed on benign surfaces~\cite{wheel}. To combine the strengths of both systems, wheel-legged robots have recently emerged as a promising hybrid platform~\cite{RollingETH}. 

However, the advantage of wheels fades at terrain discontinuities. Most wheel-legged platforms rely on stepping motions that treat the wheels as footholds~\cite{JunhoLee1,JunhoLee2,CTBC}. This approach is well-suited to systems with relatively small wheels but compromises the speed and efficiency of continuous rolling. An alternative is to roll over discontinuities by exploiting wheel traction and drive torque. With large enough wheels, a larger radius engages obstacle edges at a shallower approach angle while tire compliance absorbs impacts, so that moderate discontinuities are overcome by the wheel itself. Although early attempts have been made~\cite{BlindStair,Mobinn}, robust traversal of challenging terrain has yet to be achieved. We target this regime directly, building a wheeled bipedal robot with 20-inch wheels.

Unlike many locomotion problems that can be addressed primarily through visual perception~\cite{Hwangbo3,JunhoLee2}, rolling-based locomotion is fundamentally a \emph{contact-dominated} problem, governed by local wheel--terrain interaction. The optimal behavior at each instant is highly sensitive to this interaction, yet the decisive contact state is unobservable to the robot. A rolling wheel cannot be densely instrumented along its continuously rotating surface, and external sensing captures the surrounding terrain rather than the contact interface itself. The signal is moreover complex, as small variations can produce repeated impact, restitution, transient loss of contact, and slip~\cite{Nonsmooth}, depending not only on rigid-body geometry but also on tire compliance and internal materials~\cite{Pacejka2005}.

Prior work has approached this challenge on the modeling side, refining the simulator until contact dynamics transfer faithfully to hardware. This succeeds for point-foot contact~\cite{CSY}, but the contact of a wheel is far harder to formulate, and modeling the wheel deformation explicitly~\cite{Mobinn} has left a substantial sim-to-real gap. Therefore, rather than chasing a faithful contact model, we exploit the contact information available in simulation to learn a contact-informed latent representation, which is subsequently inferred from proprioceptive history at deployment. A similar idea has proven effective in contact-rich manipulation, where contact-related state that cannot be observed at deployment is recovered implicitly through learning~\cite{ContactRich3}.

On our platform, the large wheels do more than carry the robot over obstacles. Because contact spreads across spatially distinct locations on the wheel surface, they yield a rich signal of the wheel--terrain interaction, which we treat as privileged information during training. Accordingly, we formulate a task- and platform-specific representation of wheel--terrain multi-contact information and integrate it into Regularized Online Adaptation (ROA)~\cite{DWB}. To prepare the noisy raw contact data for distillation, we design a temporal filter and a sorting scheme that produce a refined, fixed-size contact vector.

\begin{figure*}[!th]
    \vspace{0.3cm}
    \centering
    \includegraphics[width=\linewidth]{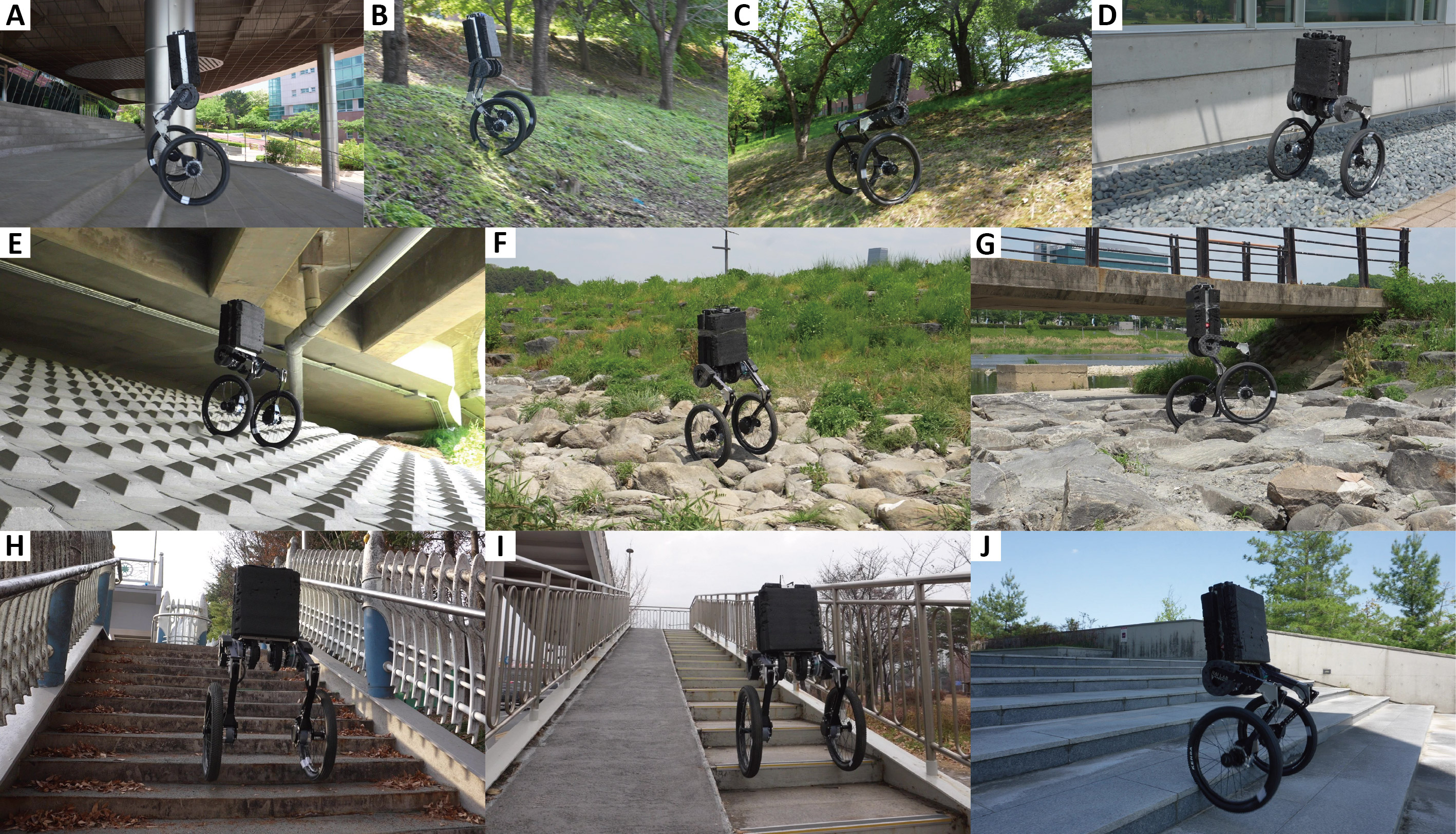}
    \setlength{\abovecaptionskip}{-0.5cm}
    \caption{\textbf{Real-world experiments across diverse challenging environments: }Representative deployments include (A) rapid stair descent, (B,C) steep-slope descent and ascent, traversal of (D) gravel bed, (E) studded slope, and (F,G) rocky terrain, and continuous stair ascent at (H) 9~cm, (I) 14~cm, and (J) 15~cm riser heights.}
    \vspace{-0.5cm}
    \label{fig:experiment}
\end{figure*}

We deploy the proposed contact-informed controller on our in-house wheeled bipedal robot across diverse outdoor environments, as shown in Fig.~\ref{fig:experiment}. The test set spans continuous contact variation at high speed on slopes and gravel, spatially unpredictable contact on studded and rocky terrain, and periodic discontinuities on staircases with riser heights up to 15~cm. Along the way, it encounters contact events such as wheel slip, whole-body lateral drift, and hard impacts, while continuing locomotion in these representative deployments. We further compare it against baselines without the contact-informed mechanism on a 16~cm indoor step ascent and an outdoor rocky traversal, where it achieves a markedly higher success rate on both. Together, these diverse deployments and ablation studies support our central premise that the large wheels make continuous rolling over irregular terrain physically feasible, while the contact-informed controller improves performance on our platform across the tested conditions.

\begin{samepage}
Our main contributions are as follows:
\begin{itemize}
    \item We develop RAIWAY, an in-house large-wheeled bipedal robot with 20-inch wheels that enables rolling locomotion across irregular and discontinuous terrain.
    \item We propose a processing pipeline that transforms simulator-derived multi-contact events around a large wheel into a structured contact-informed latent representation, which is subsequently used for online adaptation.  
    \item We demonstrate real-world rolling locomotion across diverse irregular and discontinuous terrain without external perception or contact sensing.
\end{itemize}
\end{samepage}

\section{RELATED WORK}

\subsection{Existing Wheeled Bipedal Robots and Blind Locomotion}
Wheeled bipedal robots have recently gained attention as compact platforms that combine the efficiency of wheels with the obstacle-negotiation ability of legs. Ascento introduced a compact wheeled bipedal design with four-bar-linkage legs, each constrained to a single actuated DoF and optimized primarily for hopping over obstacles~\cite{Ascento1}. Its successor, Ascento Pro, uses larger wheels and serial 2-DoF legs, on which Klemm et al.~\cite{Nonsmooth} planned discontinuity traversal with a non-smooth trajectory optimization that handles contact switches and impacts, while Chamorro et al.~\cite{BlindStair} demonstrated reinforcement-learning-based blind ascent of a 15~cm step using an asymmetric actor--critic with a stair-climbing mode flag. More recently, the TRON1 platform pursues agility by pairing small wheels with 3-DoF legs, on which Li et al.~\cite{CTBC} proposed contact-triggered blind climbing, where wheel--obstacle contact initiates a learned leg-lifting maneuver.

Across these platforms, discontinuities are mainly handled by interrupting or supplementing rolling with dedicated leg-mediated behavior. This separation between wheeled travel and leg-based clearance suits specific obstacle classes but does not extend to general terrain, leaving the wheel underused as a terrain-adaptive contact body. Large compliant wheels instead negotiate moderate discontinuities mechanically: a larger radius reduces the effective approach angle, while tire compliance absorbs impacts, maintains traction, and smooths edge engagement. Our platform therefore adopts large off-road tires, making rolling the primary traversal mode, with the legs regulating posture and load transfer.

\subsection{Explicit and Implicit Use of Privileged Information}

In reinforcement learning, leveraging privileged information during training has proven effective for enhancing performance and narrowing the sim-to-real gap~\cite{LBC, asyAC}. Privileged information has been used in two closely related ways: explicit estimation of deployable states and implicit adaptation via latent variables. 

For instance, Ji et al.~\cite{JGH} concurrently trained a locomotion policy with an estimator to predict privileged quantities. Alternatively, Kumar et al.~\cite{RMA} proposed a two-stage teacher--student pipeline that reconstructs a latent encoding of privileged factors from proprioceptive histories. Fu et al.~\cite{DWB} later proposed Regularized Online Adaptation (ROA), which jointly trains the teacher and student modules while regularizing the privileged latent to remain predictable from onboard history. Jeon et al.~\cite{JSH} applied ROA to whole-body manipulation under unknown object properties.

Contact-related quantities have also been used as privileged inputs. Lee et al.~\cite{Hwangbo2} included foot contact states and forces, while Miki et al.~\cite{Hwangbo3} further included contact normals. More recently, Wang et al.~\cite{CTS} included foot contact forces among the privileged inputs in a concurrent teacher--student framework. In these works, contact enters as one of many privileged quantities used to improve locomotion in general. Our problem, however, is itself \emph{contact-dominated}, so contact is the signal of interest rather than one input among many. It is also richer here than at a point foot, which is what the proposed representation is designed to capture.

\section{METHOD}

\subsection{RAIWAY: The Large-Wheeled Bipedal Robot}

\begin{figure}[h]
    \centering
    \setlength{\abovecaptionskip}{-0.1cm}
    \includegraphics[width=\linewidth]{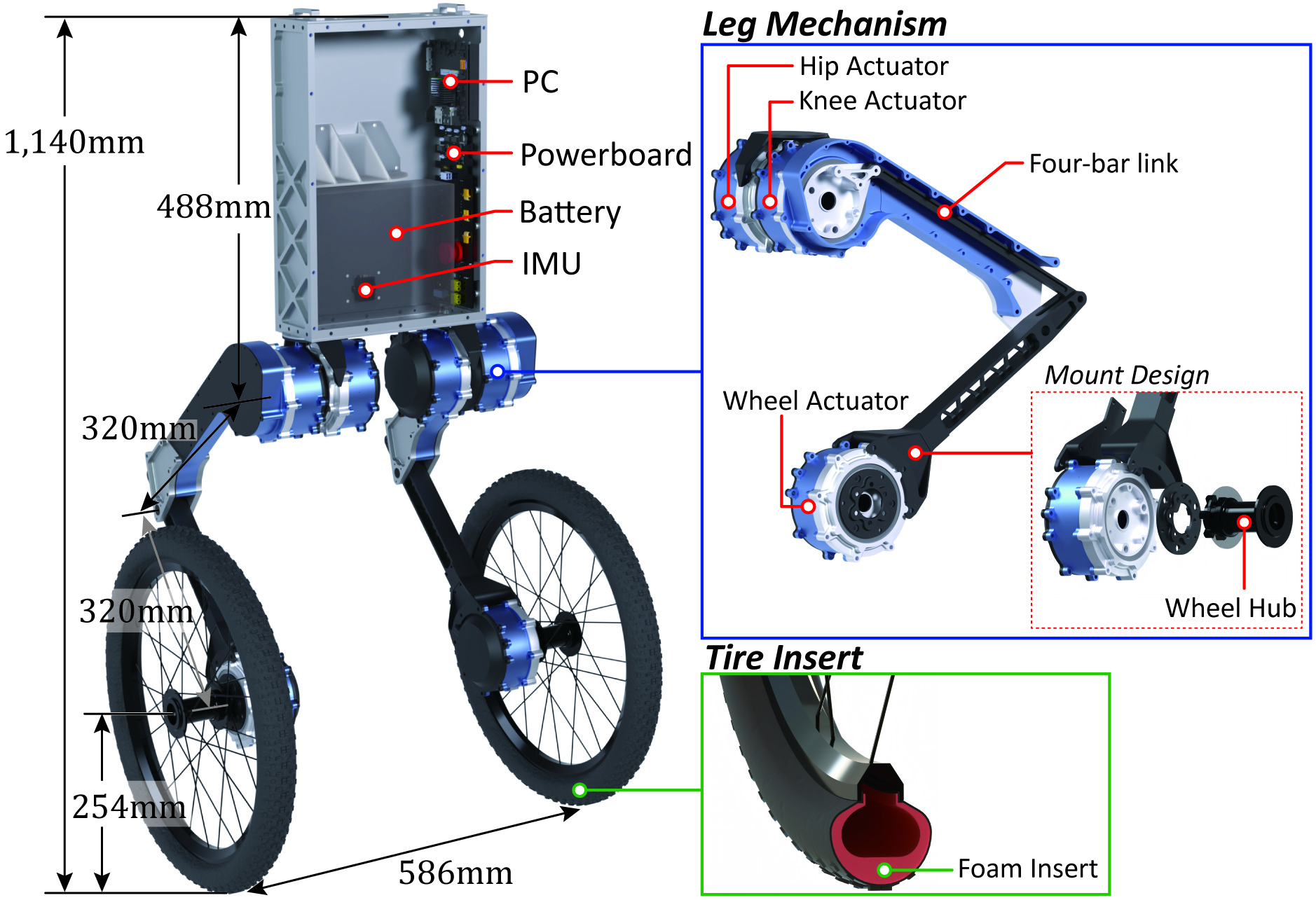}
    \caption{\textbf{Design of RAIWAY: } 
    The main view shows the overall geometry and key dimensions, with insets detailing the leg mechanism, the wheel-actuator mount, and a cross-section of the closed-cell foam tire insert.}
    \label{fig:robot_platform}
    \vspace{-0.0cm}
\end{figure}

The proposed system is a 6-DoF wheeled bipedal robot, shown in Fig.~\ref{fig:robot_platform}. Each leg comprises two pitch joints and one wheel joint, and the robot employs differential steering without hip roll or yaw joints. We reuse validated subsystems from our previous in-house quadruped platform RAIBO2~\cite{RAIBO2}---including the actuators, link mechanism, motor drivers, and onboard power system---while redesigning the body frame and leg layout for a two-legged morphology. The robot weighs approximately 28~kg including the battery and stands 1.14~m tall in its nominal configuration. It is equipped with an onboard Jetson AGX Orin computer, a MicroStrain 3DM-GX5-AR IMU, and joint encoders on all actuated joints.

Each wheel is mounted directly to the actuator output, using 20-inch off-road bicycle tires. To improve robustness, we replace pneumatic inflation with closed-cell polymer foam inserts, which provide more repeatable compliance during impacts and avoid both the excessive rebound of high tire pressure and the rim strikes of low pressure. This predictability also benefits sim-to-real transfer, as a more consistent tire response narrows the gap between the contact dynamics experienced in simulation and those encountered during deployment.

\subsection{Contact-Informed Policy Learning}

\begin{figure*}[!th]
    \vspace{0.3cm}
    \centering
    \includegraphics[width=\linewidth]{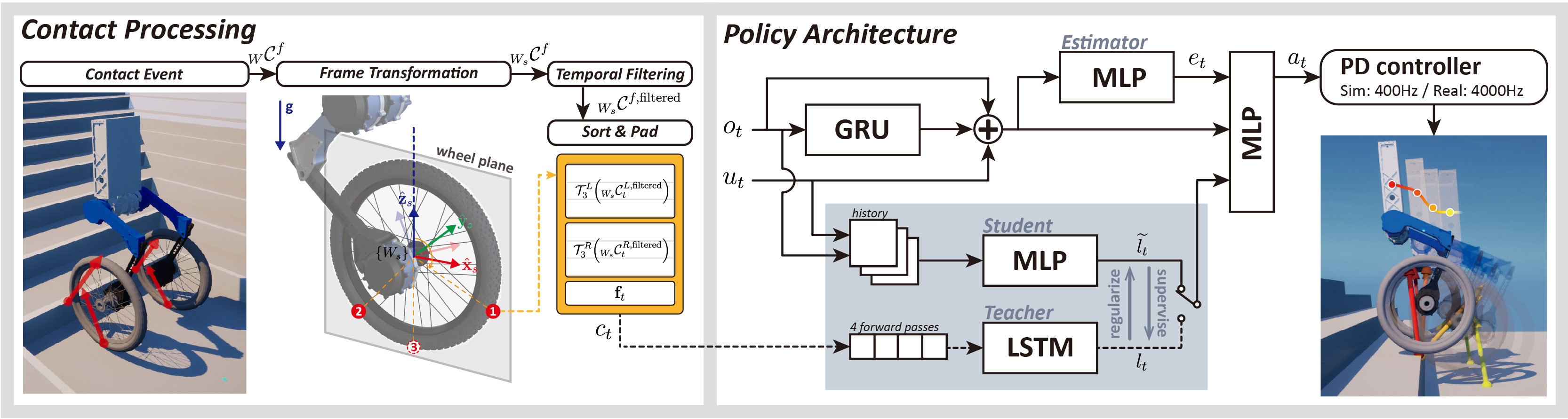}
    \setlength{\abovecaptionskip}{-0.5cm}
    \caption{\textbf{Overview of the proposed contact-informed policy learning framework: }
    The large wheels of our platform produce informative multi-contact patterns, yielding a diverse set of spatially distributed contact events during training. These noisy multi-contact events from the simulator are assembled into a structured vector $c_t$. A recurrent encoder maps this vector sequence into a latent representation, which a student module learns to infer from onboard proprioceptive history for deployment. The GRU encoder has a hidden size of $128$, the estimator and policy are $[256, 128]$ MLPs, the teacher encoder is an LSTM with a hidden size of $8$ producing the $32$-dimensional latent $\mathbf{l}_t$, and the student is a $[256, 128]$ MLP, all with LeakyReLU activations.}
    \vspace{-0.5cm}
    \label{fig:overall}
\end{figure*}

The proposed controller, shown in Fig.~\ref{fig:overall}, combines recurrent proprioceptive encoding, explicit estimation of predefined physical quantities, and contact-informed latent adaptation. At each control step, the robot receives a command $\mathbf{u}_t = [v_x^{\mathrm{des}}, \omega_z^{\mathrm{des}}]^\top$, consisting of the desired forward velocity and yaw rate. The proprioceptive observation is defined as

\begin{equation}
\mathbf{o}_t =
[\boldsymbol{\omega}_t,\; \mathbf{g}_t,\; \mathbf{q}^{\mathrm{leg}}_t,\; \dot{\mathbf{q}}_t,\; \mathbf{a}_{t-1},\; \mathbf{a}_{t-2}]^\top
\end{equation}

where \(\boldsymbol{\omega}_t\) is the base angular velocity, \(\mathbf{g}_t\) is the projected gravity vector, \(\mathbf{q}^{\mathrm{leg}}_t\) is the leg joint positions excluding the wheel angles, \(\dot{\mathbf{q}}_t\) is the joint velocities, and $\mathbf{a}_{t-1}$ and $\mathbf{a}_{t-2}$ are the two most recent actions.

A GRU encodes this observation into a recurrent feature, while a concurrent estimator predicts a compact vector $\mathbf{e}_t$ of deployable quantities informative for rolling locomotion:

\begin{equation}
\mathbf{e}_t =
\left[
\mathbf{v}_t,\;
 h_t,\;
 \mathbf{p}_{f,z,t}-\mathbf{r},\;
 \mathbf{v}_{f,y,t},\;
 \mathbf{v}_{f,xz,t}-\boldsymbol{\Omega}_t \odot \mathbf{r}
\right]^\top
\end{equation}

Here $\mathbf{v}_t$ is the base linear velocity, $h_t$ the base height, and $\mathbf{r}$ the wheel-radius vector. The remaining entries are stacked per leg: $\mathbf{p}_{f,z,t}-\mathbf{r}$ gives the wheel clearances, $\mathbf{v}_{f,y,t}$ the lateral foot velocities, and $\mathbf{v}_{f,xz,t}-\boldsymbol{\Omega}_t \odot \mathbf{r}$ the deviation from ideal rolling, which serves as a slip indicator.

At the core of the proposed framework, the controller is conditioned on a contact-informed latent representation distilled from privileged contact information, whose construction and online adaptation are detailed in the following subsections. The policy outputs desired positions for the four leg joints and desired angular velocities for the two wheel joints, tracked by a low-level PD controller with \(K_p = 50\), \(K_d = 0.5\) for the leg joints and \(K_p = 0\), \(K_d = 2.5\) for the wheel joints.

\subsection{Processing Contact Events in Simulation}

At each simulator substep, we collect the raw contact set of each wheel in the world frame \(\{W\}\), denoted by \({}_{W}\mathcal{C}^{f}\):
% \begin{equation}
% \resizebox{0.9\linewidth}{!}{$
%     \displaystyle {}_{W}\mathcal{C} = \left\{ \left( {}_{W}\mathbf{p}_{i,f}, {}_{W}\mathbf{I}_{i,f} \right) \;\middle|\; i \in \{1, \dots, n_f\}, \; f \in \{L, R\} \right\}
% $}
% \end{equation}
\begin{equation}
\resizebox{0.9\linewidth}{!}{$
    \displaystyle {}_{W}\mathcal{C}^{f} = \left\{ \left( {}_{W}\mathbf{p}_{i,f}, {}_{W}\mathbf{I}_{i,f} \right) \;\middle|\; i \in \{1, \dots, n_f\} \right\}, \qquad f \in \{L, R\}
$}
\end{equation}
where \(\mathbf{p}\) and \(\mathbf{I}\) denote the contact position and impulse, respectively, and $n_f$ is the number of recorded contacts for wheel \(f \in \{L,R\}\). The privileged contact information is not used in this raw form. Because rigid-contact simulation produces variable and temporally discontinuous wheel--terrain contact events, we convert them into a rotation-invariant, filtered, and fixed-size vector before passing them to the teacher module.

\subsubsection{Transforming into Rotation-Invariant Frame}
A wheel-centric representation suits learning better than the world frame, but the native wheel frame rotates with the wheel, making contact locations phase-dependent. To remove this ambiguity, we define a wheel-static frame $\{W_s\}$ that is invariant to wheel rotation. The $y_s$-axis is aligned with the wheel axle and is therefore normal to the wheel plane; we denote its unit vector by $\hat{\mathbf{y}}_s$. We then define the $z_s$-axis as the direction opposite to the projection of gravity onto the wheel plane:

\begin{equation}
\mathbf{z}'_s = \mathbf{g} - (\mathbf{g}^{\top}\hat{\mathbf{y}}_s)\hat{\mathbf{y}}_s,
\qquad
\hat{\mathbf{z}}_s = -\frac{\mathbf{z}'_s}{\|\mathbf{z}'_s\|}
\end{equation}

Finally, \(\hat{\mathbf{x}}_s\) is obtained from the right-hand rule. By this construction, the \(x_s\)--\(z_s\) plane coincides with the wheel plane and \(\hat{\mathbf{x}}_s\) aligns with the forward direction at the nominal posture. The origin of \(\{W_s\}\) is placed at the geometric center of the wheel's collision body, which is modeled as a cylinder. The raw contact set \({}_{W}\mathcal{C}^{f}\) is transformed accordingly into \({}_{W_s}\mathcal{C}^{f}\), after which the $x_s$ and $z_s$ components of each transformed contact position are normalized by the wheel radius, and the $y_s$ component is normalized by the half cylinder length, mapping each coordinate into $[-1, 1]$.

\vspace{0.2cm}
\subsubsection{Temporal Filtering of Contact Events} 
\begin{figure}[!t]
    \centering
    \setlength{\abovecaptionskip}{-0.2cm}
    \includegraphics[width=\linewidth]{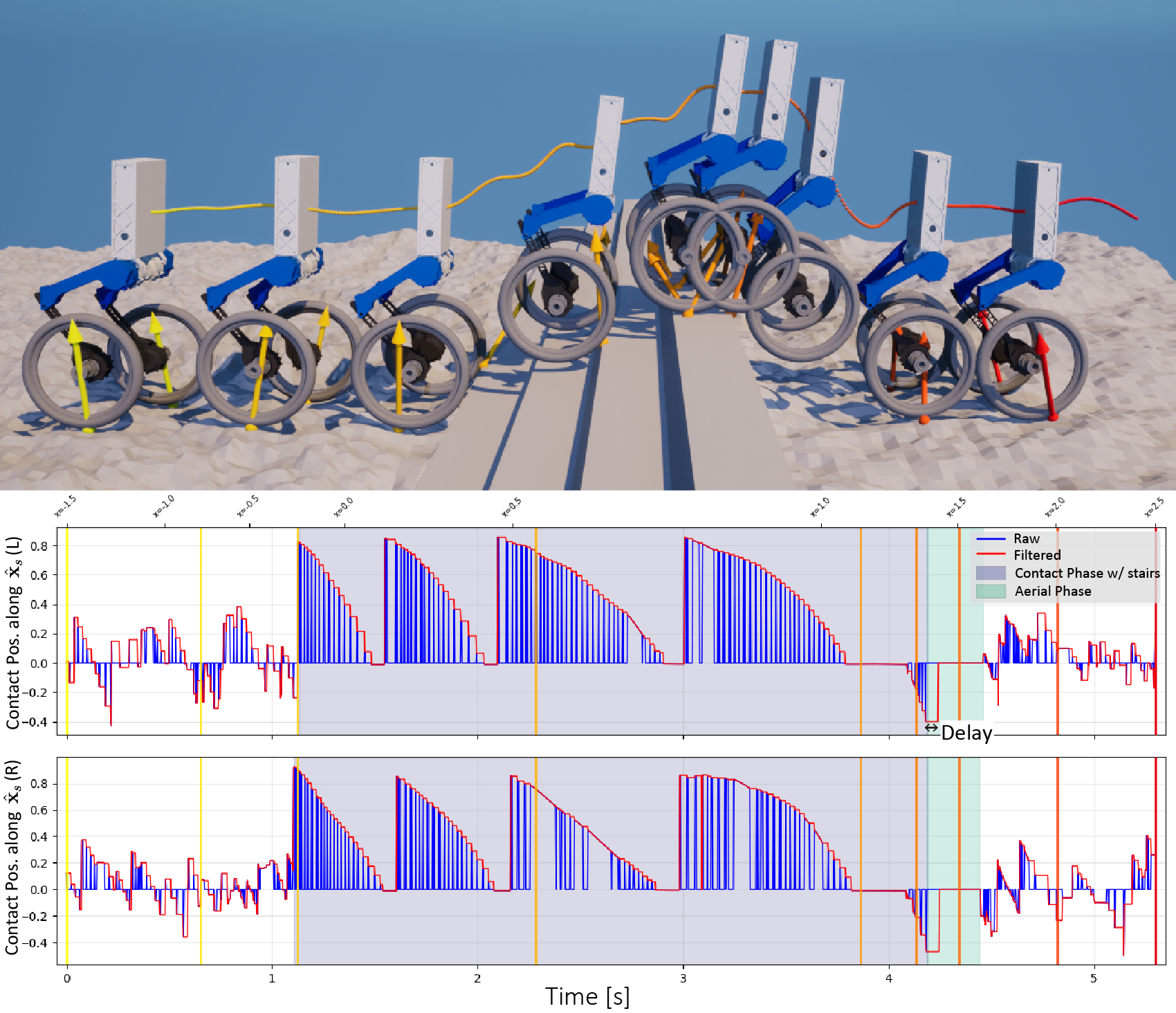}
    \caption{\textbf{Temporal filtering during a representative simulated rollout: } The robot traverses a flat--stairs--drop--flat terrain sequence, where the flat sections feature low-amplitude fractal height variations. The graphs compare the \(x_s\)-coordinate of the first angularly sorted contact position for the two wheels before and after temporal filtering. On the flat terrain, the raw signal flickers despite continuous rolling contact, whereas the filtered signal preserves variation. During stair ascent, the filtered signal captures the step-edge locations with a distinct piecewise evolution. During the drop segment, a true aerial phase is induced, and the debounce filter introduces a short delay in the contact-state transition.}
    \label{fig:filtering}
    \vspace{-0.6cm}
\end{figure}

The simulated contact set flickers even when the robot is stationary or rolling steadily. We address this with an asymmetric debounce filter: a newly detected contact is accepted within one simulator substep, whereas disappearance is committed only after 25 consecutive substeps (62.5~ms), retaining the previously accepted set during this falling-edge window. The window is based on the 99th-percentile simulated contact-loss durations on smooth terrain and a 5~cm step, where sustained contact was confirmed on hardware. A longer window would further delay genuine contact-loss detection. Fig.~\ref{fig:filtering} shows the suppression of temporal flicker. We denote the resulting set after temporal filtering by ${}_{W_s}\mathcal{C}^{\mathrm{filtered}}$.

\vspace{0.2cm}
\subsubsection{Constructing the Contact Information Vector}
To impose a consistent order and produce a fixed-size input, we define a deterministic sort-and-pad operator $\mathcal{T}_3^{f}(\cdot)$. It sorts ${}_{W_s}\mathcal{C}^{f,\mathrm{filtered}}$ by angular position around the wheel circumference in the $x_s$--$z_s$ plane and writes the elements into three slots, initialized with a default template at the wheel bottom, $x_s=y_s=0,\,z_s=-1$ with zero impulse, distinguishable from active contact. Three slots retain the expressiveness needed for simultaneous bottom, front-lower, and back-lower contacts. Across the sampled terrains, four or more contacts occurred in at most 0.305\% of wheel samples, each containing a pair within $10.6^\circ$ circumferentially, equivalent to a 4.3~mm step height. The closest pair is then repeatedly merged until exactly three remain, where each merge of contacts $i$ and $j$ produces
\begin{equation}
\mathbf{p}_{ij} = \frac{\|\mathbf{I}_i\|\mathbf{p}_i + \|\mathbf{I}_j\|\mathbf{p}_j}{\|\mathbf{I}_i\| + \|\mathbf{I}_j\|},
\qquad
\mathbf{I}_{ij} = \mathbf{I}_i + \mathbf{I}_j
\end{equation}
% \begin{equation}
% \resizebox{0.9\linewidth}{!}{$
%     \mathcal{M}\!\left((\mathbf{p}_i,\mathbf{I}_i),\,(\mathbf{p}_j,\mathbf{I}_j)\right)
%     :=
%     \left(
%     \frac{\|\mathbf{I}_i\|\,\mathbf{p}_i + \|\mathbf{I}_j\|\,\mathbf{p}_j}
%          {\|\mathbf{I}_i\| + \|\mathbf{I}_j\|},
%     \;
%     \mathbf{I}_i + \mathbf{I}_j
%     \right)
% $}
% \label{eq:merge}
% \end{equation}

Angular sorting is a natural choice for rolling over terrain discontinuities because the contact location on the wheel carries direct geometric meaning: a contact near the bottom of the wheel indicates rolling along a flat surface, whereas one displaced toward the front of the wheel indicates engagement with an obstacle edge. This ordering ensures that recurring configurations consistently occupy the same slots, allowing the encoder to associate each slot with a fixed geometric role.

The full contact information vector is then assembled by concatenating the per-wheel outputs of $\mathcal{T}_3^{f}(\cdot)$ with a wheel-level privileged feature block $\mathbf{f}_t$:
\begin{equation}
\mathbf{c}_t
=
\begin{bmatrix}
\mathcal{T}_3^{L}\!\left({}_{W_s}\mathcal{C}_{t}^{L,\mathrm{filtered}}\right) &
\mathcal{T}_3^{R}\!\left({}_{W_s}\mathcal{C}_{t}^{R,\mathrm{filtered}}\right) &
\mathbf{f}_t
\end{bmatrix}
\end{equation}
where $\mathbf{f}_t$ is defined as
\vspace{-0.1cm}
\begin{equation}
\mathbf{f}_t =
\left[
\mathbf{p}_{f,z,t}-\mathbf{r},\; 
\mathbf{v}_{f,t},\;
 \mathbf{v}_{f,xz,t}-\boldsymbol{\Omega}_t \odot \mathbf{r}
\right]^\top
\end{equation}

% \vspace{0.2cm}
\subsubsection{Contact-Informed Latent Adaptation}

\begin{table}[!t]
\vspace{0.3cm}
\centering
\caption{Contact-Related Domain Randomization Ranges}
\vspace{-0.3cm}
\label{tab:domain_randomization}
\renewcommand{\arraystretch}{1.2}
\resizebox{\linewidth}{!}{
    \begin{tabular}{
        p{0.29\linewidth}
        p{0.41\linewidth}
        p{0.41\linewidth}
    }
    \toprule
    Category & Parameter & Range / Unit \\
    \midrule

    Wheel Collision Body
        & Radius
        & $+\,\mathcal{U}(-4,\,0)~[\mathrm{cm}]$ \\
    & Thickness
        & $+\,\mathcal{U}(-4,\,0)~[\mathrm{cm}]$ \\
    & Center offset
        & $+\,\mathcal{U}(-2,\,2)~[\mathrm{cm}]$ \\
    & Mass
        & $+\,\mathcal{U}(-0.05,\,0.20)~[\mathrm{kg}]$ \\
    & Inertia
        & $+\,\mathcal{U}(-0.02,\,0.03)
        ~[\mathrm{kg\!\cdot\!m^2}]$ \\

    \midrule

    Terrain Parameters
        & Restitution coefficient
        & $\mathcal{U}(0.2,\,0.3)$ \\
    & Restitution threshold
        & $\mathcal{U}(0.5,\,0.8)~[\mathrm{m/s}]$ \\

    \midrule

    Wheel Slip
        & Random prob. in episode
        & $0.1~[\%]$ \\
    & Contact-based prob. in episode
        & $\mathrm{clip}
        \!\left(1.25\,\mathbf{p}_{z,\max}+1,\,0,\,1\right)$ \\
    & Friction coefficient
        & $\mathcal{U}(0.1,\,0.4)$ \\

    \bottomrule
    \end{tabular}
}
\vspace{-0.4cm}
\end{table}
\begin{table}[!t] %H
\caption{Reward Functions}
\vspace{-0.3cm}
\label{table:rew}
\centering
\renewcommand{\arraystretch}{1.2}
\setlength{\tabcolsep}{6pt}
\resizebox{\linewidth}{!}{%
    \begin{tabular}{|c|l|c|}
    \hline
    \textbf{Reward} & \multicolumn{1}{c|}{\textbf{Expression}} & \textbf{Coeff.} \\
    \hline
    Forward Velocity Tracking &
    $\exp\!\big(-({v}_{x}^{des} - {v_x}')^2\big),\quad {v_x}'=\operatorname{sign}(v_x)\,v_{xz}$ & $1.5$ \\
    \hline
    Lateral Velocity Tracking &
    $\exp\!\big(-0.5{v_y}^2\big)$ & $1.0$ \\
    \hline
    Angular Velocity Tracking &
    $\exp\!\big(-1.5({\omega}_{z}^{des}-\omega_z)^2\big)$ & $1.0$ \\
    \hline
    Orientation &
    $R_{zz}$ & $1.0$ \\
    \hline
      Foot Slip &
  $\sum_{i \in \{L,R\}} \mathbf{1}[i \text{ in contact}]\,\big|\,v_{f,xz,i} - r_i\,|\Omega_i|\,\big|$ & $-0.5$ \\
    % Foot Slip &
    % $\sum_{i \in \{L,R\}} \big(\lVert v_{f,xz,i} \rVert - \lVert \Omega_i r_i \rVert\big)$ & $-0.5$ \\
    \hline
    Contact & 
      $\sum_{i \in \{L,R\}} \mathbf{1}[i \text{ not in contact}]$ & $-0.1$  \\ 
    \hline
    Base Motion &
    $\omega_x^2+\omega_y^2$ & $-2.0$ \\
    \hline
    Base Height &
    $(h - h^{\mathrm{des}})^2$ & $-40.0$ \\
    \hline
    Joint Position 1$^{\dagger}$ &
    $\|(\mathbf{q} - \mathbf{q}_{\mathrm{nominal}})\odot \mathbf{w}_q\|$ & $-1.0$ \\
    \hline
    Joint Position 2 &
    $(p_{f,x,L} - p_{f,x,R})^2$ & $-0.5$ \\
    \hline
    Joint Velocity$^{\dagger}$ &
    $\|\dot{\mathbf{q}} \odot \mathbf{w}_{\dot q}\|$ & $-5.0\times 10^{-4}$ \\
    \hline
    Joint Torque &
    $\|\boldsymbol{\tau}\|^2$ & $-1.0\times 10^{-4}$ \\
    \hline
    Action Smoothness 1$^{\dagger}$ &
    $\|(\mathbf{a}_{t}-\mathbf{a}_{t-1})\odot \mathbf{w}_s\|$ & $-0.8$ \\
    \hline
    Action Smoothness 2$^{\dagger}$ &
    $\|(\mathbf{a}_{t}-2\mathbf{a}_{t-1}+\mathbf{a}_{t-2})\odot \mathbf{w}_s\|$ & $-0.4$ \\
    \hline
    Torque Smoothness 1 &
    $\|\boldsymbol{\tau}_t-\boldsymbol{\tau}_{t-1}\|^2$ & $-2.0\times 10^{-7}$ \\
    \hline
    Torque Smoothness 2 &
    $\|\boldsymbol{\tau}_t-2\boldsymbol{\tau}_{t-1}+\boldsymbol{\tau}_{t-2}\|^2$ & $-1.0\times 10^{-7}$ \\
    \hline
    \end{tabular}%
}
\par\vspace{0.05cm}
\parbox{\linewidth}{\scriptsize $^{\dagger}$Weighting vectors applied to balance the position- and velocity-commanded joints: $\mathbf{w}_q=[1,\allowbreak 1,\allowbreak 0,\allowbreak 1,\allowbreak 1,\allowbreak 0]$, $\mathbf{w}_{\dot q}=[1,\allowbreak 1,\allowbreak 0.1,\allowbreak 1,\allowbreak 1,\allowbreak 0.1]$, $\mathbf{w}_s=[1,\allowbreak 1,\allowbreak 0.05,\allowbreak 1,\allowbreak 1,\allowbreak 0.05]$.}
\vspace{-0.4cm}
\end{table}

During training, contact information is available at the 400~Hz simulator substep rate, four times the 100~Hz policy rate, so four contact vectors are collected per control step. The teacher module, an LSTM, is unrolled over these substeps and outputs a latent \(\mathbf{l}_t\) comprising the four hidden states. The adaptation module, an MLP, receives the most recent \(N=20\) observations and commands, \(\{(\mathbf{o}_{\tau}, \mathbf{u}_{\tau})\}_{\tau=t-N+1}^{t}\), and predicts \(\tilde{\mathbf{l}}_t\). The two modules are trained jointly through
\begin{equation} 
L_{\mathrm{adaptation}} =
\lambda \|\mathbf{l} -\mathrm{sg}\![\tilde{\mathbf{l}}]\|_2^2
+\|\mathrm{sg}\![\mathbf{l}] -\tilde{\mathbf{l}}\|_2^2
\end{equation}
where \(\mathrm{sg}[\cdot]\) denotes the stop-gradient operator. The coefficient \(\lambda\) regularizes the teacher latent to remain predictable from the student, and is annealed from \(0\) to \(1\) over training. The actor is conditioned on \(\mathbf{l}_t\) for most of training, and in a final phase the contact encoder is frozen and the actor is switched to \(\tilde{\mathbf{l}}_t\), so that the policy is fine-tuned on the imperfectly recovered latent it will actually receive at deployment.

\subsection{Training Details}

We train the policy with PPO \cite{PPO} in Raisim \cite{Raisim} using 200 parallel environments for 60{,}000 iterations, on an AMD Ryzen 5995WX CPU and an NVIDIA GeForce RTX 3080 Ti GPU in approximately 42 hours.

We apply domain randomization for sim-to-real transfer, with ranges in Table~\ref{tab:domain_randomization}. Extending Lee et al.~\cite{JunhoLee2}, who randomized the wheel radius, we randomize various collision-body parameters to better represent actual tire dynamics, and randomize terrain parameters to reflect the slip and rebound observed in real experiments. Discontinuous features such as steps and stairs are generated with heights up to 15~cm. The reward design reflects our primary objective of continuous rolling driven by wheel rotation rather than stepping: we omit the gait-shaping and foot-clearance rewards typical of legged locomotion, and instead apply a contact penalty when a wheel loses ground contact and a slip term to discourage excessive slip, as listed in Table~\ref{table:rew}.

\section{RESULTS}

We first analyze the contact-vector inputs and the learned latent in simulation (Section~\ref{IV-A}). We then evaluate the system through real-world ablation studies (Section~\ref{IV-B}), outdoor deployments (Section~\ref{IV-C}), and failure-case analysis (Section~\ref{IV-D}).

\subsection{Analysis of Contact-Vector and Contact-Informed Latent} \label{IV-A}

We ablate the components of $\mathbf{c}_t$ and the number of contact slots per wheel in simulation, comparing against \textbf{\textit{AC + estimator}}, which follows~\cite{JGH} and corresponds to \textbf{\textit{Ours}} without the contact-informed latent. Each variant is trained once from a single random seed under identical settings, differing only in $\mathbf{c}_t$. Each controller runs 1{,}000 trials on a 15~cm staircase, and Table~\ref{tab:contact_ablation} reports the mean percentage of the route traversed before falling.

Removing contact positions or impulses reduces traversal distance from 74.4\% to 50.4\% and 53.6\%, respectively. Removing wheel-level features, which partly overlap with the concurrent estimator targets, has a smaller effect, yielding 64.1\%. Using one or two slots also substantially degrades performance. Among the tested variants, only those retaining both positions and impulses in three slots outperform \textit{AC + estimator}.

\begin{figure}[!t]
    \centering
    \setlength{\abovecaptionskip}{-0.2cm}
    \includegraphics[width=\linewidth]{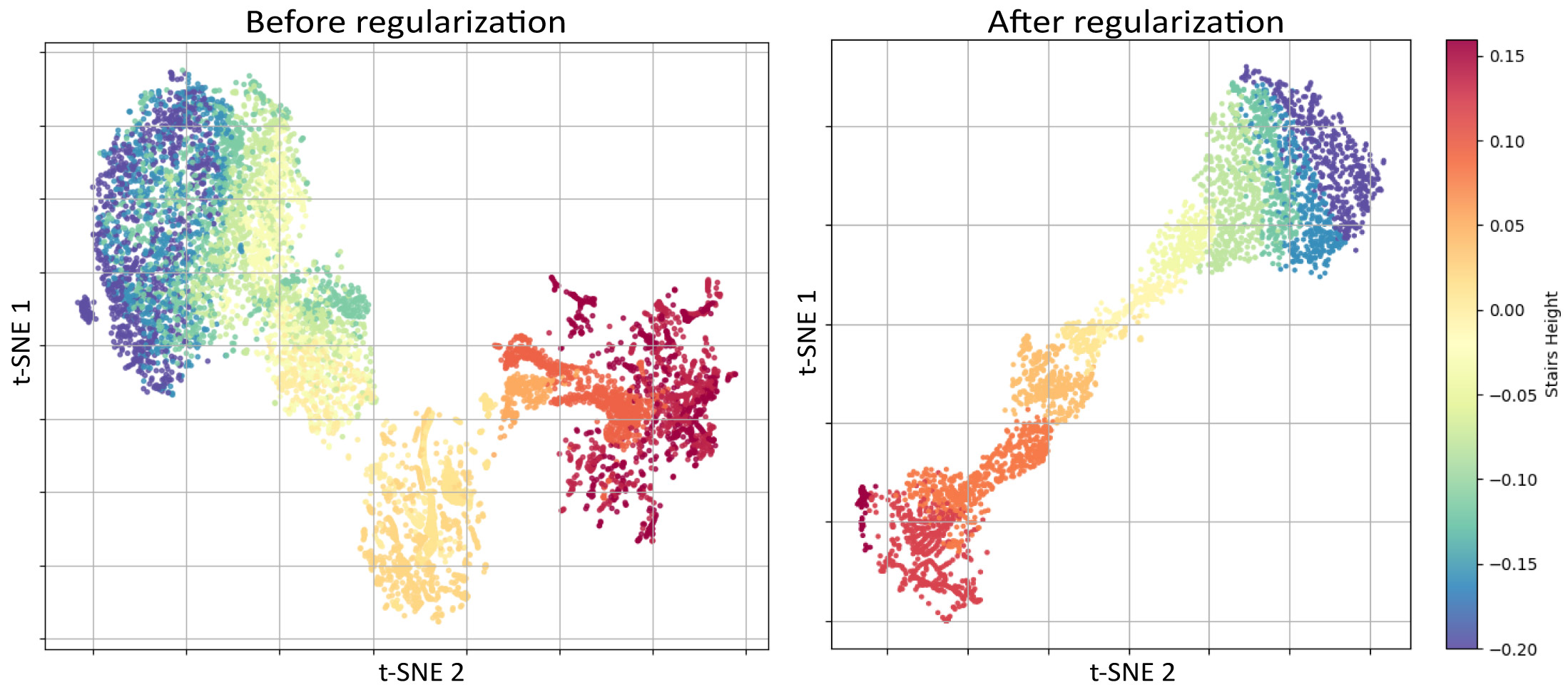}
    \caption{\textbf{t-SNE projection of $\mathbf{l}_t$.} The left panel shows the latent at mid-training before the regularization term was activated ($\lambda = 0$), and the right panel shows the latent at the end of training ($\lambda = 1$).}
    \label{fig:tsne}
    \vspace{-0.6cm}
\end{figure}

We further examine the physical information reflected in the teacher latent $\mathbf{l}_t$ using t-SNE and Integrated Gradients (IG). For t-SNE, staircase rollouts span riser heights from $-20$ to $+15$~cm, with negative values denoting descent. Fig.~\ref{fig:tsne} shows ordering by riser height at mid-training before regularization, becoming smoother and more continuous by training completion with full regularization. For IG, we analyze 16{,}000 states from smooth terrain, steps, pipes, and stairs, sampled toward equal representation across contact regimes. Using the mean normalized input as reference, we aggregate attributions across latent dimensions with the $\ell_2$-norm. Contact position, impulse, wheel clearance, wheel velocity, and slip account for approximately 50.5\%, 13.8\%, 7.7\%, 22.5\%, and 5.6\% of total block-level attribution, respectively.

% \begin{table}[t]
%     \centering
%     \caption{Contact-vector ablation in simulation.}
%     \label{tab:contact_ablation}
%     \footnotesize
%     \setlength{\tabcolsep}{3pt}
%     \begin{tabular*}{\columnwidth}{@{\extracolsep{\fill}}lrlr@{}}
%         \toprule
%         Variant & Dist. [\%] & Variant & Dist. [\%] \\
%         \midrule
%         Ours & \textbf{74.4} & 2 slots & 46.2 \\
%         w/o wheel-level features & 64.1 & 1 slot & 44.2 \\
%         w/o contact position & 50.4 & w/o contact pos \& imp & 46.5 \\
%         w/o contact impulse & 53.6 & AC + estimator & 57.2 \\
%         \bottomrule
%     \end{tabular*}
% \end{table}

\begin{table}[!t]
\vspace{0.3cm}
\centering
\caption{Contact-vector ablation in simulation}
\label{tab:contact_ablation}
\footnotesize
\renewcommand{\arraystretch}{1.2}
\setlength{\tabcolsep}{6pt}
\vspace{-0.2cm}
\resizebox{\linewidth}{!}{%
\begin{tabular}{l | c | l | c}
\hline\hline
\textbf{Variant} &
\makecell{\textbf{Traversal}\\\textbf{Dist. [\%]}} &
\textbf{Variant} &
\makecell{\textbf{Traversal}\\\textbf{Dist. [\%]}} \\
\hline
\textbf{Ours}           & \textbf{74.4} & w/o contact position   & 50.4 \\
w/o wheel-level features & 64.1        & 2 slots                & 46.2 \\
AC + estimator           & 57.2        & 1 slot                 & 44.2 \\
w/o contact impulse      & 53.6        & w/o contact pos \& imp & 42.9 \\
\hline\hline
\end{tabular}%
}
\vspace{-0.5cm}
\end{table}

\subsection{Ablation Study} \label{IV-B}

We compare the following controllers, all sharing the same observation, network architecture, reward, and domain randomization:

\begin{samepage}
\begin{itemize}
    \item \textbf{\textit{Plain AC}}: Plain asymmetric actor--critic.
    \item \textbf{\textit{AC + estimator}}: \textit{Plain AC} augmented with a concurrent estimator, following~\cite{JGH}.
    \item \textbf{\textit{Ours w/o Estimator}}: Ours w/o concurrent estimator.
    \item \textbf{\textit{Ours w/o Filtering}}: Ours w/o temporal filtering.
\end{itemize}
\end{samepage}

The ablation study consists of two representative deployment tasks. Experiment 1 targets a 16~cm step, corresponding to roughly 63\% of the wheel radius. Each controller is evaluated over 20 trials: \emph{Fail} if the robot topples; \emph{Unstable} if a completed ascent has a base yaw deviation above 0.5~rad or a peak pitch angular velocity $|\omega_y|$ above 1.5~rad/s; and \emph{Success} otherwise. Experiment 2 evaluates sustained traversal over distributed irregular contacts, a 20~m round-trip on the same rocky terrain shown in Fig.~\ref{fig:experiment}(F). The traversal distance is measured at ten checkpoints along the round-trip route, reported as the percentage of the route completed, and averaged across 5 trials. Results are reported in Table~\ref{tab:ablation_results}.

\textit{Plain AC} and \textit{AC + estimator}, without the contact-informed mechanism, struggle across both experiments but in opposite ways. \textit{AC + estimator} improves step climbing but performs worse on rocky terrain. A post hoc simulation analysis showed larger errors on rocky terrain for eight of ten estimator outputs, including all six wheel-specific quantities. Such diverse contacts are hard to capture with a few scalars, so the policy that acts directly on these estimates inherits their inaccuracy.

In contrast, \textit{Ours} and its reduced variants exceed both baselines across the two experiments. Comparing \textit{AC + estimator} and \textit{Ours w/o Estimator}, which inject privileged information through the concurrent estimator and the contact-informed latent respectively, rocky traversal rises from 39\% to 83\%, identifying the latent as the larger contributor. Adding the concurrent estimator on top of the latent, as in \textit{Ours}, achieves a 100\% step success rate and the highest rocky traversal distance. The two mechanisms are thus complementary: the latent handles the distributed contacts of irregular terrain, and the estimator secures the low-variability single discontinuities where its scalar predictions are reliable. \textit{Ours w/o Filtering} falls short of \textit{Ours} but still outperforms both baselines, suggesting that the filter refines the latent rather than providing its core benefit.

\begin{table}[!t]
\vspace{0.3cm}
\centering
\caption{Ablation study results}
\label{tab:ablation_results}
\footnotesize
\renewcommand{\arraystretch}{1.2}
\setlength{\tabcolsep}{6pt}
\vspace{-0.2cm}
\resizebox{\linewidth}{!}{%
\begin{tabular}{l | c c | c c c | c}
\hline\hline
& \multicolumn{2}{c|}{\textbf{Components}} & \multicolumn{3}{c|}{\textbf{Experiment 1}} & \textbf{Experiment 2} \\
\cline{2-3} \cline{4-6} \cline{7-7}
\textbf{Controller} & \makecell{\textbf{Contact-}\\\textbf{informed}} & \textbf{Estimator} & \textbf{Fail} & \textbf{Unstable} & \textbf{Success} & \makecell{\textbf{Traversal Dist. [\%], }\\\textbf{mean\,[min, max]}} \\
\hline
Plain AC             & X & X & 6 & 9 & 5  & 69\,[35, 100] \\
AC + estimator       & X & O & 4 & 2 & 14 & 39\,[5, 65] \\
Ours w/o Estimator   & O & X & 2 & 4 & 14 & 83\,[25, 100] \\
Ours w/o Filtering   & O & O & 2 & 1 & 17 & 81\,[25, 100] \\
\textbf{Ours}        & O & O & \textbf{0} & \textbf{0} & \textbf{20} & \textbf{95\,[75, 100]} \\
\hline\hline
\end{tabular}
}
\vspace{-0.5cm}
\end{table}

\begin{figure}[!t]
    \centering
    \setlength{\abovecaptionskip}{-0.2cm}
    \includegraphics[width=\linewidth]{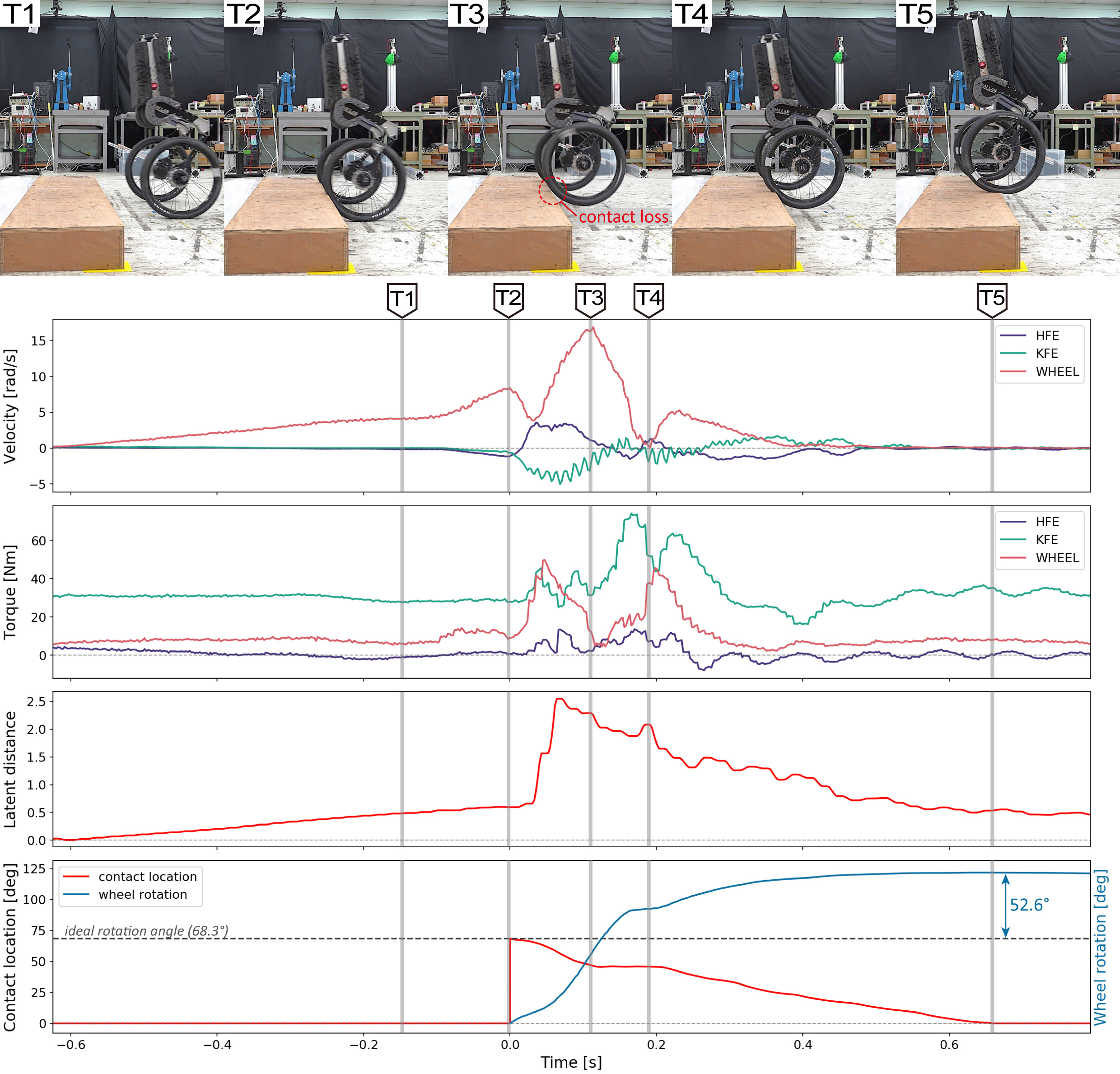}
    \caption{\textbf{Temporal response during a 16~cm step ascent: } Velocities and torques of the left leg joints, together with the latent distance $d_t$, aligned to the ascent onset ($t = 0$). The last panel shows the representative contact location and wheel rotation, with the dashed horizontal line indicating the geometrically required wheel rotation for the 16~cm step ascent.}
    \label{fig:response}
    \vspace{-0.6cm}
\end{figure}

Fig.~\ref{fig:response} further illustrates the real-robot response during a 16~cm step ascent with the proposed controller. We define $d_t$ as the Euclidean distance between the deployment-time latent $\tilde{\mathbf{l}}_t$ and the mean latent collected while standing still, measuring departure from the standing contact reference. At the initial step-edge impact (snapshot~T2), the wheel velocity drops and $d_t$ rises sharply. Rebound then causes a brief contact loss around T3, during which the wheel rotates rapidly and the joint velocities peak. After re-engagement, the contact location progressively moves toward the bottom of the wheel as the ascent proceeds, while $d_t$ trends back toward the reference value. By T5, the accumulated wheel rotation exceeds the $68.3^\circ$ geometrically required to pivot over the step by $52.6^\circ$, indicating additional wheel rotation associated with the transient contact loss.

\subsection{Outdoor Deployment across Diverse Terrains} \label{IV-C}

\begin{figure*}[!t]
    \centering
    \subfloat[Clearing a rock near the wheel radius]{%
        \includegraphics[width=0.49\linewidth]{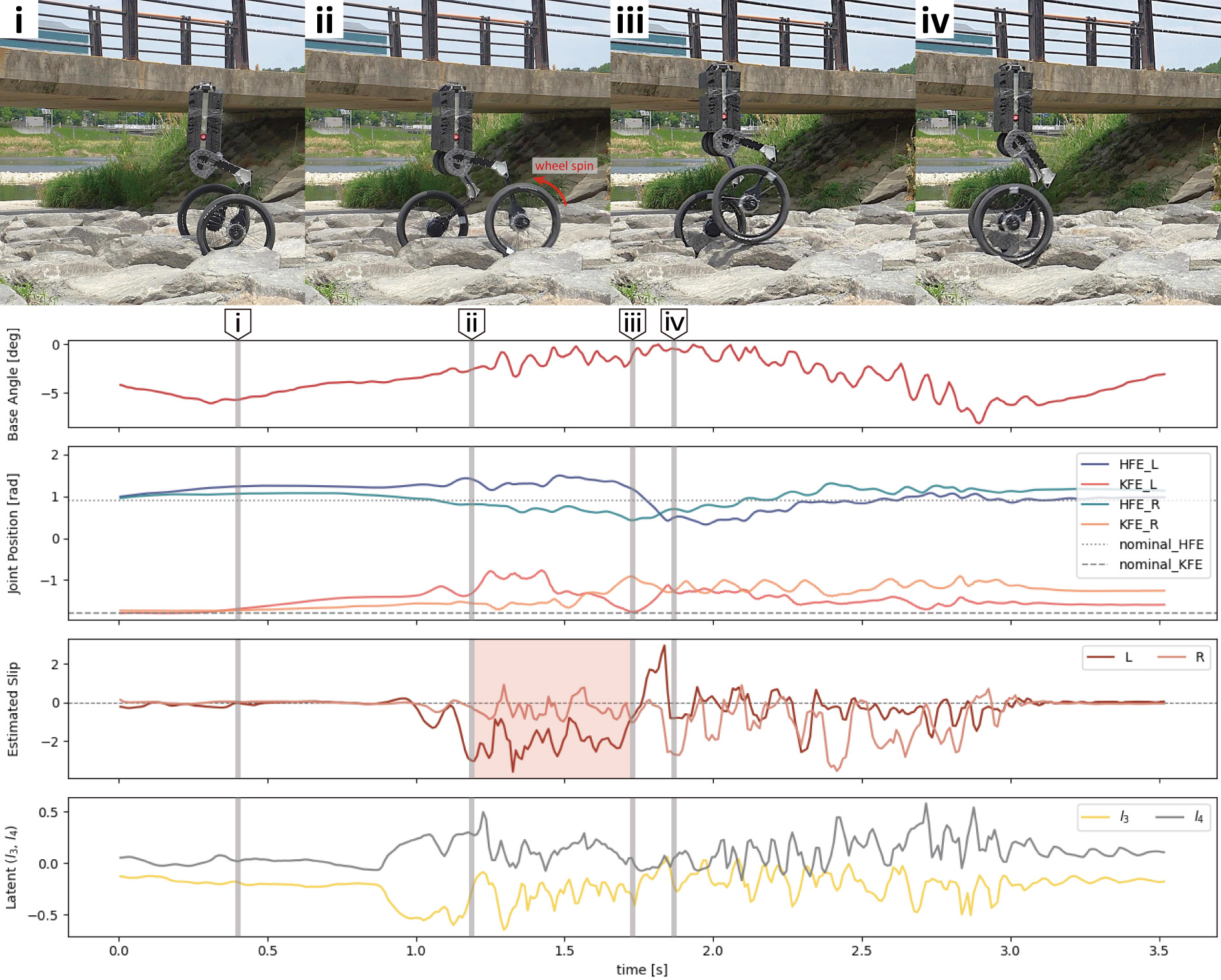}%
        \label{fig:real_data1}%
    }\hfill
    \subfloat[Whole-body lateral drift on a studded slope]{%
        \includegraphics[width=0.49\linewidth]{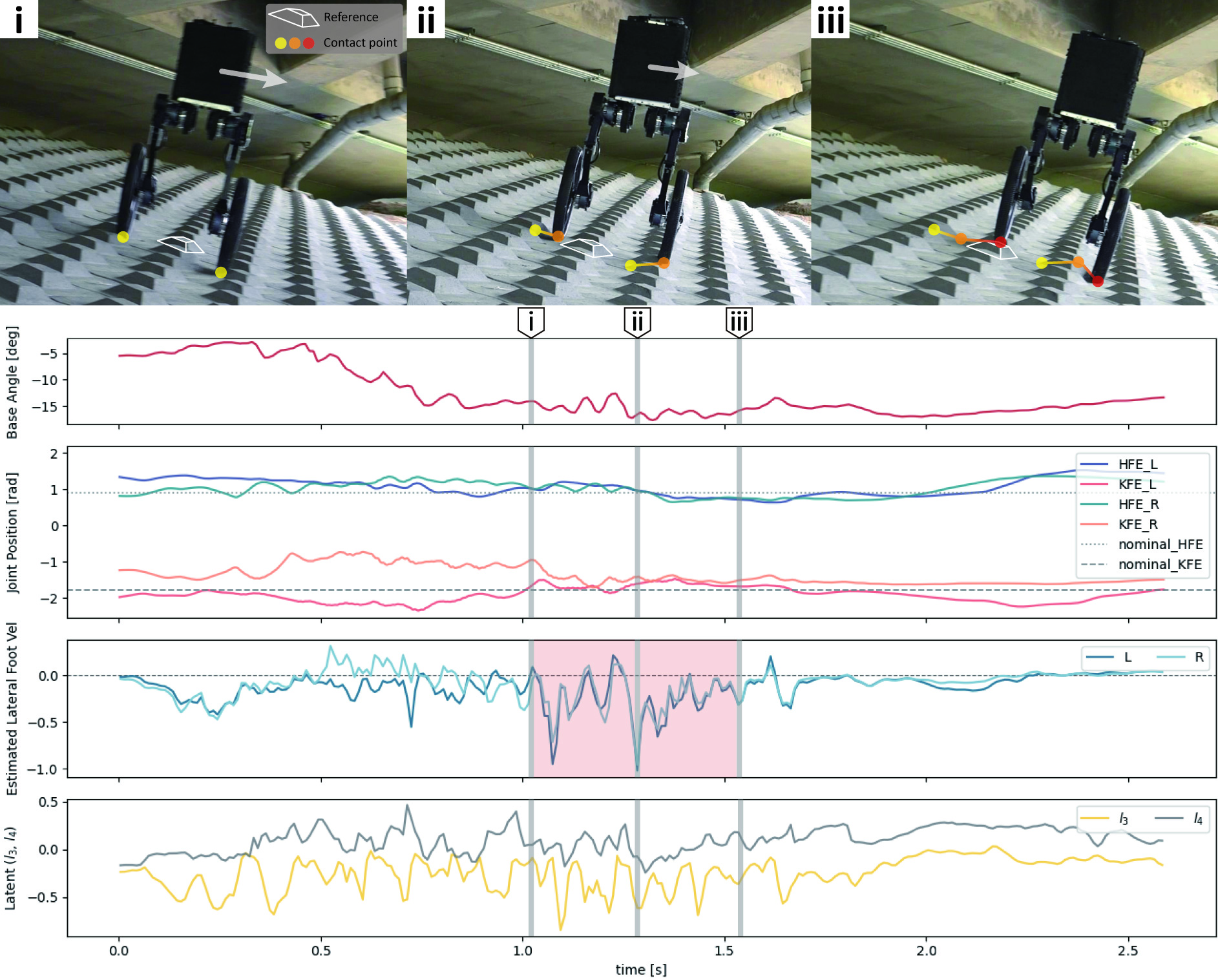}%
        \label{fig:real_data2}%
    }
    \setlength{\abovecaptionskip}{2pt}
    \caption{\textbf{Responses to abrupt contact events during outdoor deployment:}
    Two representative episodes involving challenging contact conditions. The plots show base angle, leg-joint positions, the event-relevant estimator output, and two latent components ($l_3,\,l_4$).
    (a) The left wheel spins against a wheel-radius-scale rock in (i)--(ii), after which the left leg rolls over the rock on its wheel in (iii), producing a pronounced body oscillation after landing in (iv).
    (b) The studded slope shifts the wheel--terrain contact locations laterally.}
    \label{fig:robustness}
    \vspace{-0.4cm}
\end{figure*}

We deployed the learned policy across a range of outdoor environments shown in Fig.~\ref{fig:experiment}, with dimensions and command speeds summarized in Table~\ref{tab:envs}. A single policy was used across all environments without retuning, terrain-specific switching, or external perception. The environments fall into three groups, each probing a different aspect of contact-dominated rolling.

The first group (Fig.~\ref{fig:experiment}~A--D) involves terrain without severe discontinuities, allowing traversal at high speed while the contact conditions vary throughout. The rapid stair descent (A) subjects the wheel to repeated impacts, the steep slopes (B,C) shift the contact normal and traction demand with the slope angle, and the gravel bed (D) yields and shifts under the wheel. Across all four environments, the controller maintained continuous rolling at high commanded speeds without falls.

\begin{table}[!t]
\centering
\caption{Summary of real-world environments}
\label{tab:envs}
\footnotesize
\renewcommand{\arraystretch}{1.2}
\setlength{\tabcolsep}{4pt}
\vspace{-0.2cm}
\resizebox{\linewidth}{!}{%
\begin{tabular}{c l l c}
\toprule
\textbf{Label} & \textbf{Environment} & \textbf{Dimensions} & \textbf{Command [m/s]} \\
\midrule
A & Stair descent        & 18~cm (H) $\times$ 30~cm (W), 21 steps     & 3.0 \\
B & Downhill slope       & 40$^\circ$ incline                         & 2.0 \\
C & Uphill slope         & 20$^\circ$ incline                         & 2.5 \\
D & Gravel bed           & 6--8~cm stones                             & 3.0 \\
E & Studded slope        & 20$^\circ$ incline, studs of 5~cm height   & 1.5 \\
F & Rocky terrain        & $\leq$22~cm (H), 25--50~cm (W)             & 1.0$^{\dagger}$ \\
G & Rocky terrain        & $\leq$22~cm (H), 50--80~cm (W)             & 1.0$^{\dagger}$ \\
H & Stair ascent         & 9~cm (H) $\times$ 30~cm (W), 20 steps      & 0.51$^{\ast}$ \\
I & Stair ascent         & 14~cm (H) $\times$ 76~cm (W), 20 steps     & 0.28$^{\ast}$ \\
J & Stair ascent         & 15~cm (H) $\times$ 80~cm (W), 10 steps     & 0.42$^{\ast}$ \\
\bottomrule
\end{tabular}%
}
\vspace{2pt}\\
{\footnotesize\raggedright $^{\dagger}$Operator modulated the command below this value in response to terrain conditions.\newline $^{\ast}$Measured mean traversal speed.\par}
\vspace{-0.6cm}
\end{table}

The second group (Fig.~\ref{fig:experiment}~E--G) targets diverse and spatially unpredictable contacts. On the studded slope (E) and the rocky terrain (F,G), the contact location on the wheel changes abruptly depending on which protrusion or rock face the wheel happens to engage, and cannot be anticipated from the overall terrain profile. Fig.~\ref{fig:robustness} shows these local interactions in the estimator and latent traces. In the rock-clearing case, the left-wheel slip estimate turns negative during wheel spin before~(ii) and positive between~(ii) and~(iii) during clearance, while the latent components separate slightly earlier than this explicit response. In the studded-slope case, the lateral body motion appears as a sustained lateral-foot-velocity estimate. The estimator and latent thus remain responsive during abrupt, short-duration contact changes.

The final group (Fig.~\ref{fig:experiment}~H--J) involves periodic discontinuities. At every riser, the wheel must engage the edge at a contact point displaced far from the bottom of the wheel, convert traction into upward motion without excessive slip, and re-establish rolling on the tread before the next edge arrives; under comparable edge and surface conditions, taller risers impose greater geometric demands. On the 9~cm staircase (H), six trials yielded a mean ascent speed of 0.51~m/s with a peak of 0.62~m/s. On the 14~cm staircase (I), the robot completed 11 consecutive round trips with a mean ascent speed of 0.28~m/s, reaching a peak of 0.38~m/s. On the 15~cm staircase (J), the robot succeeded in 3 of 10 trials, with a mean ascent speed of 0.42~m/s in the successful trials.

\subsection{Failure Cases} \label{IV-D}

Two failure modes were observed during deployment. First, the robot succeeded in all 20 trials on the indoor 16~cm wooden step, but on the ten-step 15~cm granite staircase (Fig.~\ref{fig:experiment}~J) it reached the top in only 3 of 10 trials (85.4\% per step) despite the lower riser. The wooden step's sharp edge gives the tire tread blocks a corner to catch on; the granite staircase, with rounded edges and a rough, uneven surface, offers none, so both wheels intermittently slipped and the robot pitched forward and toppled. Our rigid-contact simulator models neither such edge and surface variation nor the tread engagement and compliance of the real tire, leaving this contact mismatch unresolved.

Second, strong wheel rebound occasionally caused high-frequency oscillations in the actions and joint torques. We attribute this to wheel-restitution mismatch: the measured tire restitution is about 0.61, but training sampled only up to 0.3 because larger values made the policy overly conservative. Consequently, rebound events on hardware can expose the policy to observation sequences outside the training distribution. The post-landing oscillation after (iv) in Fig.~\ref{fig:robustness}(a) is one such response that decayed harmlessly. In trials where the oscillation persisted, it propagated into unfavorable whole-body motion that ended in a fall. Addressing these failure modes will require training over more diverse edge and surface geometries, along with better contact modeling or fine-tuning on real-world data.

\section{CONCLUSION}
We presented RAIWAY, a large-wheeled bipedal robot and a contact-informed control approach for blind rolling locomotion over irregular and discontinuous terrain. The framework distills simulator-derived multi-contact events into a latent representation that the deployed controller infers from proprioceptive history, enabling contact-responsive locomotion without external perception or contact sensing. Across the tested conditions, the resulting system traversed steep slopes, gravel beds, studded surfaces, and rocky terrain, and demonstrated continuous ascent on the tested outdoor staircases with riser heights up to 15~cm, spanning a range of outdoor terrain and discontinuous contact conditions. The benefits of contact-informed control were demonstrated on our large-wheeled platform, where obstacle traversal is driven primarily by wheel rolling.

Future work will focus on incorporating exteroceptive perception so that anticipatory terrain awareness and the present contact-responsive behavior reinforce each other, enabling traversal at higher speeds over more complex discontinuities.

%%%%%%%%%%%%%%%%%%%%%%%%%%%%%%%%%%%%%%%%%%%%%%%%%%%%%%%%%%%%%%%%%%%%%%%%%%%%%%%%

\bibliographystyle{IEEEtran}
\bibliography{IEEEabrv,ref}
\end{document}